\documentclass[onecolumn]{article}
\usepackage{bagrow}

\usepackage{subfigure}
\usepackage{soul}
\usepackage{multirow}
\usepackage{enumitem}

\newcommand{\phat}{\hat{p}}
\newcommand{\pline}{\overline{p}}
\newcommand{\ptilde}{\Tilde{p}}

\newcommand{\xhat}{\hat{x}}

\newcommand{\xtilde}{\Tilde{x}}

\definecolor{darkspringgreen}{rgb}{.1725, .6274, .1725}

\makeatletter
\if@twocolumn
\fi
\makeatother
\author[1,2]{Abigail Hotaling}
\author[1,2,*]{James Bagrow}
\affil[1]{Department of Mathematics \& Statistics, University of Vermont, Burlington, VT, United States}
\affil[2]{Vermont Complex Systems Center, University of Vermont, Burlington, VT, United States}
\affil[*]{\corrauthinfo{james.bagrow@uvm.edu}{bagrow.com}}

\title{Accurate inference of crowdsourcing properties when using efficient allocation strategies}
\date{April 27, 2022}

\begin{document}

\maketitle
\begin{abstract}
\singlespacing
Allocation strategies improve the efficiency of crowdsourcing by decreasing the work needed to complete individual tasks accurately.
However, these algorithms introduce bias by preferentially allocating workers onto easy tasks, leading to sets of completed tasks that are no longer representative of all tasks.
This bias challenges inference of problem-wide properties such as typical task difficulty or crowd properties such as worker completion times, important information that goes beyond the crowd responses themselves.
Here we study inference about problem properties when using an allocation algorithm to improve crowd efficiency.
We introduce Decision-Explicit Probability Sampling (DEPS), a novel method to perform inference of problem properties while accounting for the potential bias introduced by an allocation strategy.
Experiments on real and synthetic crowdsourcing data show that DEPS outperforms baseline inference methods while still leveraging the efficiency gains of the allocation method.
The ability to perform accurate inference of general properties when using non-representative data allows crowdsourcers to extract more knowledge out of a given crowdsourced dataset.
\end{abstract}


\section{Introduction}

Crowdsourcing has become a valuable source of information for a wide variety of applications~\cite{howe2006rise,brabham2008crowdsourcing,kittur2013future}, particularly in areas where humans can perform work not possible or not well suited to computational methods. 
Common crowdsourcing tasks include generating labeled training data for machine learning algorithms~\cite{snow2008cheap,yan2011active,kamar2012combining}, performing text recognition or natural language tasks~\cite{snow2008cheap}, completing surveys~\cite{behrend2011viability}, or generating novel questions or other creative inputs~\cite{bongard2013crowdsourcing,salganik2015wiki,mcandrew2017reply,wagy2017crowdsourcing,berenberg2018efficient}.
However, using human volunteers or paid crowd workers brings with it a host of new problems~\cite{kittur2013future,garcia2016challenges}.
Human workers are often more costly than computational methods, receiving responses from workers can be relatively slow, and one must worry about worker motivation~\cite{zheng2011task,kaufmann2011more} and reliability~\cite{eickhoff2013increasing,kittur2013future}.
Even with reliable workers, challenging tasks often require a consensus approach: multiple workers are given a task and their set of (possibly noisy) responses are aggregated to generate the final response for the crowdsourcer~\cite{sheng2008get}.

Algorithmic crowdsourcing focuses on computational approaches to dealing with issues introduced when using a crowdsourcing solution~\cite{karger2011iterative,karger2014budget,oyama2013accurate,li2016crowdsourcing}.
These algorithms improve the crowd's efficiency by enabling more accurate information gain, and often reduce the numbers of workers needed.
Some methods learn the reliability of workers and minimize or eliminate the contributions of unreliable workers~\cite{dawid1979maximum,karger2011iterative,eickhoff2013increasing}.
Allocation methods, on the other hand, focus on the tasks being crowdsourced, and attempt to allocate workers towards those tasks that can be completed most quickly and accurately~\cite{karger2014budget,li2016crowdsourcing}.
Quickly identifying those tasks where workers tend to give the same response can prevent redundant worker responses and allows the crowdsourcer to decide upon the aggregate response more efficiently.

The focus of crowd allocation algorithms has been on maximizing accuracy when aggregating responses from the crowd while minimizing the number of responses needed in order to meet budget constraints.
However, other aspects of a crowdsourcing problem are important to understand besides the final responses to the set of tasks comprising that problem.
For example, a crowdsourcer may wish to identify experts within the crowd or understand how many tasks are difficult for workers to complete compared with tasks that are easy for workers.
Or a crowdsourcer concerned about latency may want to understand how much time it takes workers to perform a task~\cite{dow2012shepherding,berenberg2018efficient}. 
Worker completion times are an example of the more general behavioral traces of workers~\cite{rzeszotarski2011instrumenting}, and a crowdsourcer may be interested in understanding the pattern of worker dynamics as they perform work.
Yet in all these cases, an allocation algorithm focused on completing tasks accurately can change the set of tasks being shown to workers, introducing a bias where easily completed tasks are more likely to receive responses from workers and leaving the crowdsourcer with a dataset that may be unrepresentative of the typical tasks or typical worker behaviour within the crowdsourcing problem.

In this work, we ask if it is possible to reason about properties of crowdsourcing problems while also employing an efficient allocation algorithm.
Of course, one can provide an unbiased view of tasks and workers by assigning problems at random, but doing so sacrifices the efficiency gains an allocation strategy can provide; it may simply not be worth the added cost to learn these problem properties, especially if developing accurate responses is the main goal.
We introduce \emph{Decision-Explicit Probability Sampling} (DEPS), a strategy to perform inference on problem properties that places the decision process of an efficient allocation algorithm within the inference model. 
DEPS can provide inference for  a variety of problem properties, although our focus here is on inferring the distribution of problem difficulty, information a crowdsourcer often desires.
Experiments with real crowdsourcing data and synthetic models show that DEPS provides accurate information about the crowdsourcing problem while using an allocation algorithm, allowing a crowdsourcer to leverage the efficiency gains of an allocation algorithm while retaining more of the problem information that can be lost when using an allocation algorithm.

The rest of this paper is organized as follows.
In Sec.~\ref{sec:background} we describe a model for a crowdsourcing problem with a fixed budget, discuss the properties of crowdsourcing problems, and detail allocations algorithms that have been introduced to improve crowdsourcing efficiency.
Section~\ref{sec:allocationmethodsbias} illustrates the challenge of property inference when using efficient allocation strategies.
Section~\ref{sec:deps} introduces Decision-Explicit Probability Sampling (DEPS), a strategy for integrating the results of an allocation algorithm into property inference.
Then, in Secs.~\ref{sec:methods} and \ref{sec:results} we describe and present experiments using DEPS on real and synthetic crowdsourcing data.
We discuss our results and future work in Sec.~\ref{sec:discussion} and conclude in Sec.~\ref{sec:conclusion}

\section{Background}
\label{sec:background}

Here we describe the model of a crowdsourcing problem along
with various problem properties that are meaningful to study for such a problem. 
We also provide background on efficient allocation algorithms a crowdsourcer can use to maximize the information gained from the crowd. 
Crowdsourcing problem models can be used to generate synthetic crowdsourced data (as a generative model), but the primary use of the model is to build inference procedures to be used when processing real crowdsourced data (as a statistical model).

\subsection{Crowdsourcing model} 
\label{subsec:courdsourcingmodel}

We employ the Dawid-Skene model for a crowdsourcing problem~\cite{dawid1979maximum}.
A crowdsourcing problem consists of $N$ tasks, each of which is considered a binary labeling task. 
Binary tasks can represent image classification tasks, survey questions, and other problems.
While relatively simplistic, binary labeling forms the basis of most crowdsourcing models.
In our results, we analyze three such tasks: recognizing whether one written statement \textit{entails} another statement, classifying a photograph of a bird as containing one species or another, and identifying whether a given web page would be relevant to a given topic (see Sec.~\ref{subsec:datasets}).
The binary task model can also generalize to categorical tasks, but such tasks can always be binarized by treating the most common individual response as a `1' and all other responses as `0'.
Let $z_i \in \{0,1\}$ be the true (unknown) label for task $i$.
A total of $M$ workers are given one or more of these tasks and respond by providing a label for the given task.
Let $y_{ij} \in \{0,1\}$ be the response of worker $j \in [1,M]$ to task $i \in [1,N]$, $a_i$ the total number of 1-labels for task $i$, $b_i$ the total number of 0-labels, and $n_i=a_i + b_i $ the total number of responses to $i$.
Individual worker responses are taken as iid for a given task. %
Not all workers necessarily respond to every task, so define $J_i$ as the set of workers who responded to task $i$, and $\left| \cup_{i=1}^N J_i \right| = M$ (we assume every worker responds to at least one task and no worker responds to the same task more than once).
Likewise, let $I_j$ be the set of tasks that worker $j$ responded to, with
$\lvert \cup_{j=1}^M I_j \rvert = N$.
The crowdsourcer can then aggregate the individual responses to reach a consensus response.
The goal is to infer best estimates $\hat{z}_i \approx z_i$ given worker responses $y_{ij}$. 
Often the crowdsourcer must approach this goal under budget constraints, as costs are incurred when employing a crowd. 
Let $T$ be a crowdsourcer's total budget, meaning the crowdsourcer can request at most $T$ individual worker responses.
This constraint leads to $\sum_i n_i \leq T$.
In seminal work, Dawid and Skene use an EM-algorithm to infer $z_i$ that outperforms the basic majority-vote strategy~\cite{dawid1979maximum} and much subsequent work has studied and generalized this approach~\cite{snow2008cheap,sheng2008get}.

\subsection{Problem properties}
\label{subsec:problemproperties}

Learning labels accurately is the goal of the crowdsourced labeling problem described above.
However, there are potentially many other properties of such problems that are of interest to researchers.
For example, what constitutes a typical task within the problem? Do tasks tend to be easy for workers to complete or are tasks difficult?
Are there distinguishing features of tasks, such as groups or categories of tasks that are similar in some way?
Likewise, a crowdsourcer may wish to learn properties of the crowd workers. 
Are there experts among the workers? Are workers reliable or not? 
Will workers tend to finish an assigned task quickly or will they require a lot of time to perform the work? 
Can we learn about worker behavioral traces~\cite{rzeszotarski2011instrumenting}, the patterns of activities the workers undertake as they respond to a task?

In general, learning both task properties and crowd properties can be of significant benefit to a crowdsourcer: tasks can be better designed accounting for features of the tasks and how workers interact with them,
costs can be better forecast and planned, and more reliable data can be generated by the crowd.
Of course, the details and importance of these properties will be heavily dependent on the type of crowdsourcing problem, but learning properties can potentially reveal significant information for the crowdsourcer, beyond that of the task labels themselves.

\subsection{Efficient allocation methods}
\label{subsec:allocalgs}

Various allocation strategies have been developed with the goal of assigning tasks to workers in order to maximize information gained about labels while minimizing costs~\cite{li2016crowdsourcing,chen2013optimistic}. 
Here we provide background on how these algorithms assign tasks. 

Requallo~\cite{li2016crowdsourcing} is a flexible allocation framework that addresses the challenge of efficient crowdsourcing for labeling tasks.
With Requallo, crowdsourcers (1) define completion requirements that tasks must satisfy to be deemed completed, (2) apply a task allocation strategy to maximize the number of completed tasks within a given budget $T$ (number of worker responses).
When a task $i$ is completed the crowdsourcer will cease seeking new responses for $i$ and be able to decide the final estimated label $\hat{z}_i$.
There are many ways to define a completion requirement.  
For example, a ratio requirement for a binary labeling task $i$ would determine that $i$ is complete when one label is assigned by workers sufficiently more often than the other label:
a ratio requirement of $c$ determines task $i$ to be complete if either $a_i / b_i > c$ or $b_i / a_i > c$, where $a_i$ and $b_i$ are the number of 1-labels and 0-labels given by workers for task $i$, respectively. 
(In practice, the Requallo authors use smoothed counts $a_i+1$ and $b_i+1$ for their requirement.)
Using the requirement, let $C(t) \subseteq [1,\ldots,N] $ be the set of completed tasks after receiving $t \leq T$ total responses from workers.
The Requallo requirement implements an early stopping rule while ensuring there is sufficient information to decide whether $z_i=0$ or $z_i=1$ with some degree of accuracy, with uncompleted tasks being undecided.

With a completeness requirement in place, Requallo then allocates tasks to workers using a Markov Decision Process (MDP) designed to quickly identify and distribute to workers those tasks which can be completed and to identify and avoid distributing those tasks which are unlikely to be completed.
Due to computational complexity, the Requallo framework determines a policy for allocating tasks using a one-step look-ahead greedy method; see Li \emph{et al.}~\cite{li2016crowdsourcing} for full details of the Requallo MDP reward function and optimization method.
Requallo's completeness requirement combined with its allocation strategy gives significant efficiency gains, with more tasks completed accurately using a fixed budget of worker responses than other allocation methods.

However, the completeness and allocation components of Requallo also introduce bias in the collected data.
Crucially, Requallo stops allocating tasks that reach completeness and the greedy MDP optimization can lead to ``hard'' tasks (those unlikely to reach completeness) being ignored in favor of ``easy'' tasks (those likely to reach completeness). 
(Note that a hard task in terms of completeness does not guarantee it is difficult or time-consuming for a particular worker to submit their response, only that the set of workers given that task will tend to disagree with one another. However, it is plausible that the same aspects of the task that lead to disagreement among workers may also affect the difficulty of the task or the time needed to submit a response.)
These properties together greatly contribute to the bias introduced by Requallo, which may be especially harmful when a crowdsourcer wants to study properties of tasks and/or workers beyond the resulting labels themselves.

Not all allocation methods explicitly distinguish decided and undecided tasks. Opt-KG~\cite{chen2013optimistic}, for example, works to optimize accuracy among the set of tasks by allocating to the appropriate tasks in order to optimize the accuracy across all tasks. 
Once the algorithm has determined a sufficiently accurate label for easier tasks, it directs the budget towards tasks that are less accurate (hard tasks).
This equates to a similar bias as the Requallo framework, where less information is obtained about easy tasks, even without an explicit decision process for those tasks.
Of course, any such allocation method can always be augmented with a decision process by including a completion requirement, and deciding if tasks are completed based on that requirement.

\section{Allocation methods lead to non-representative sets of tasks}
\label{sec:allocationmethodsbias}

Before introducing our method for inferring problem-wide properties of a crowdsourcing problem, we first illustrate the challenge introduced when a crowdsourcer employs an efficient allocation algorithm.
Of course, this challenge can be eliminated by forgoing use of an allocation algorithm, but doing so may be too cost prohibitive to consider in practice.

We applied the Requallo allocation algorithm to a crowdsourcing problem of $N=1000$ tasks averaged over 100 independent simulations defined as follows.
Each synthetic task $i$ is represented by a Bernoulli RV parameterized by $p_i$, the probability of a 1-label.
Here, tasks closer to $p_i=1/2$ are more difficult than tasks far from $1/2$, because a task with $p \approx 1/2$ will take many worker responses before we can accurately conclude whether $z_i = 1$ or $0$.
We took the prior distribution $\Pr(p_i)$ of $p_i$ to be uniform, $p_i \sim U[0,1]$. %
Knowing the true distribution of task difficulty $\Pr(p)$ in this situation allows us to compare with the observed distribution %
developed by employing the Requallo Algorithm with completeness ratio requirement $c=4$ (Sec.~\ref{subsec:allocalgs}).

\begin{figure*}
    \centering
    \subfigure[Distribution of $p$ for completed tasks]{
    \includegraphics[scale=0.40,trim=0 0 490 0,clip=true]{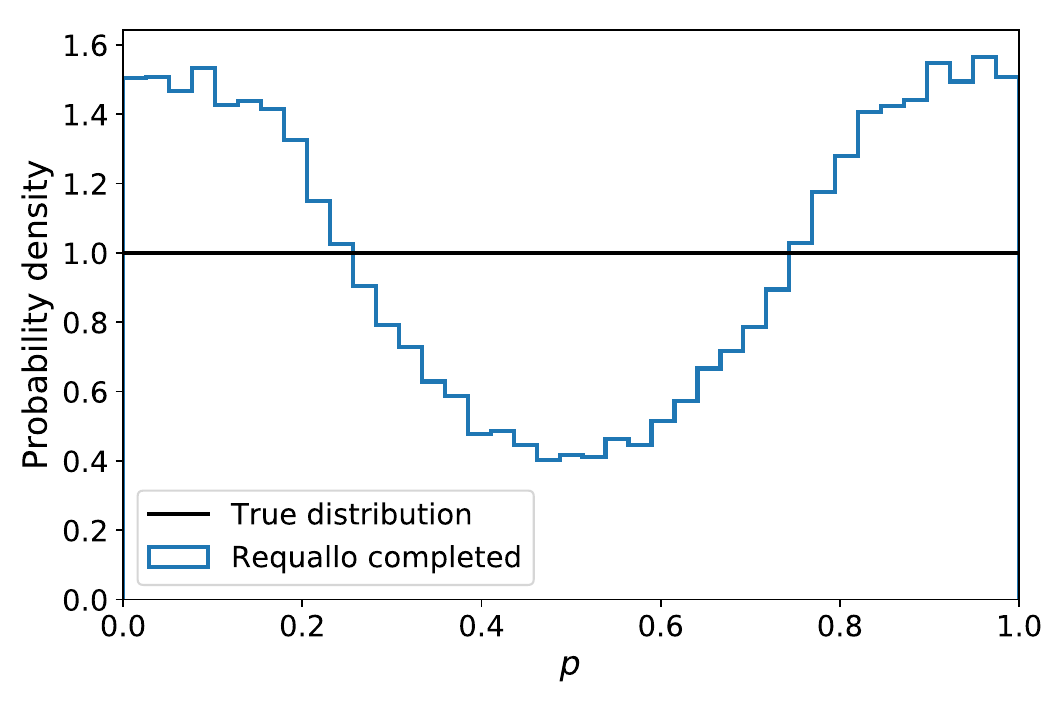}
    \label{fig:req_bias:complete_tasks_req}
}
\subfigure[Distribution of $\phat$ for completed and uncompleted tasks]{
\includegraphics[scale=0.40,trim=500 0 0 0,clip=true]{bias_uniform2.pdf}
    \label{fig:req_bias:all_tasks_req}
    }
    \caption{Requallo Allocation leads to non-representative sets of completed tasks, and biased information about problem properties.
\lett{a} Easy tasks (with $p$ far from 1/2) are over-represented in the set of completed tasks while hard tasks (with $p$ near 1/2) are under-represented.
\lett{b} Estimates of $p$ are biased, when looking at easy tasks there are pileups at $p=(0,1)$ and $p=(0.2, 0.8)$ for the "smoothed" distribution, looking at hard tasks, there are pileups at $p=0.5$.
    \label{fig:req_bias}}
\end{figure*}

Figure~\ref{fig:req_bias} compares the results a crowdsourcer using the Requallo allocation algorithm would receive with the underlying distribution of the problem.
In Fig.~\ref{fig:req_bias:complete_tasks_req} we show the difficulties of tasks deemed by Requallo to be completed versus all tasks: 
$\Pr\left(p_i \mid i \in C \right)$ 
versus 
$\Pr(p_i)$.
There is a significant bias in these distributions with Requallo-completed tasks containing far more easy tasks and far fewer difficult tasks than the overall distribution would imply.
Of course, this is an example of Requallo working by design: a net consequence of efficient allocation (and in particular the early stopping rule introduced by the completeness requirement) will always be more easy tasks than hard tasks completed. 
Yet it is evident that we are unable to capture the underlying uniform distribution with this method.

Similarly, in Fig.~\ref{fig:req_bias:all_tasks_req} we examine the distributions of estimates of problem difficulty $\phat$ for both completed and uncompleted tasks.
Examining $\Pr(\phat)$ alongside $\Pr(p)$ is important as a crowdsourcer in practice will not have access to the true parameters $p_i$ and must instead infer them from the worker responses.
Further, a crowdsourcer must often deal with small-data issues as $n_i$ may be small for a given task $i$. An example of this is when $a_i=0$ or $a_i=n_i$, leading to $\phat=0$ or $\phat=1$;
many inference procedures can fail when no counts for a given category are observed.
A common solution to this problem is to use a \textbf{\emph{smoothed}} estimate of $\phat$, where $a_i$, $b_i$, and $n_i$ are replaced with $a_i+\epsilon$, $b_i +\epsilon$,  $n_i + 2\epsilon$, respectively (we use $\epsilon=1$ as did the Requallo authors \cite{li2016crowdsourcing}). 
Examining the distributions in Fig.~\ref{fig:req_bias:all_tasks_req}, we see a significant bias in the smoothed $\Pr(\phat)$, with a trimodal sample despite the true underlying distribution being uniform. There is a ``pileup'' of values at $\phat = 1/2$ alongside pileups at $\phat = 1/2 \pm \frac{c-1}{2(c+1)}$, where $\frac{c-1}{2(c+1)}$ parameterize the Requallo requirements.
The central pileup corresponds to tasks underexplored by the crowd because Requallo deemed them too difficult and instead allocated workers to other tasks. Those other tasks, meanwhile, tend to pile up at $1/2 \pm \frac{c-1}{2(c+1)}$ because those ratios correspond to the point where Requallo decides the task is complete and ceases allocating workers to the task, preventing us from determining if the parameter $p$ is either more or less extreme than the estimate $\phat{}$. 
These phenomena show how Requallo is actually working well in terms of avoiding redundant worker responses, but also emphasizes the bias introduced by Requallo: hard tasks are underexplored, but easy tasks are also underexplored.

\section{Decision-Explicit Probability Sampling (DEPS)}
\label{sec:deps}

Suppose we wish to learn about the properties $x$ of a crowdsourcing problem, such as the distribution of problem difficulty $p$.
Accurately estimating the distribution of $p$ may be helpful for understanding the types of labeling tasks we ask workers to perform, for categorizing tasks into different groups, and for forecasting budgetary expenses such as total cost to perform typical tasks or total time we expect to wait for workers to complete their assignments.
Given sufficient data, the distribution of $p$ can be estimated relatively quickly.
Let $n_i$ be the total number of responses to binary labeling task $i$ and $a_i$ be the number of $1$-labels given by workers.
Then $\phat_i = a_i/n_i$ is the MLE (point) estimate for $p_i$.
Aggregating many estimates $\phat$ can then be used to infer the distribution $P\left(p \mid \{y_{ij}\}\right) \approx P(\phat)$.

Unfortunately, this estimation strategy will be data-intensive in practice, leading to costly crowdsourcing simply to infer the overall distribution of $p$.
Efficient algorithms are commonly introduced to ``speed up'' crowdsourcing by allowing more accurate labels to be determined with fewer worker responses.
However, as we saw in Sec.~\ref{sec:allocationmethodsbias} and Fig.~\ref{fig:req_bias}, these methods work at the expense of information about $p$: by focusing crowd resources on the easiest-to-complete tasks, the crowdsourcer is left with incomplete and unrepresentative data about the overall distribution of, in this case, $p$.
We ask in this work if it is possible to overcome the bias introduced by an efficient allocation algorithm, allowing one to both generate labels efficiently and still perform accurate inference of problem-wide features.

We propose Decision-Explicit Probability Sampling (DEPS), a method to perform inference of the problem distribution for a property $x$ when using crowdsourced data gathered by an allocation method. 
The allocation method will introduce some form of bias into the estimates $\xhat$, so our main focus is introducing a means to reason about the unbiased distribution of $x$ given the observed, biased distribution of $\xhat$.
DEPS works by explicitly incorporating the decision process of the efficient algorithm into the inferential model.

DEPS proceeds in two phases. The first is developing an inferential model for $x$ that incorporates the allocation method's decisions and the resulting bias, leading to a corrected variable $\xtilde$. Next, a probability model is fit to $\xtilde$ to infer the distribution $P(x)$ of property $x$. 
To begin, notice that a completed task $i$ may be decided to have label $\hat{z}_i = 0$ or label $\hat{z}_i=1$, depending on the $\{y_{ij} \mid j \in J_i\}$, otherwise a task is undecided.
Let the decision indicator $d_i(t)$ be the crowdsourcer's decision for task $i$ after receiving a total of $t \leq T$ responses (across all tasks):
\begin{equation}
\label{eq:d_i}
d_i(t) = \begin{cases}
      -1 & \text{if $i$ decided }z_i = 0, \\
      0 & \text{if $i$ undecided,} \\
      1 & \text{if $i$ decided }z_i = 1,
     \end{cases}
\end{equation}
with the set of completed tasks $C(t) = \{i \in [1,N] \mid d_i(t) \neq 0\}$.
DEPS incorporates the decision variable into our inference model as follows. 
For each $x_i$, we determine a posterior distribution using %
\begin{equation}
    P(x_i \mid \{y_{ij}\}, d_i) \propto \prod_{j}P(y_{ij} \mid x_i)P(x_i \mid d_i).
\label{eq:posteior_model}
\end{equation}
We then incorporate the decision $d_i$ into the prior $P(x_i \mid d_i)$ for $x_i$, choosing a different prior distribution for each decision status. 
In other words, we use the prior distributions $x_i \mid d_i \sim D_{d_i}(\Phi_{d_i})$, where $D_d$ is the prior distribution associated with decision $d$ and parameterized by $\Phi_{d}$.
Depending on the decision of task $i$ and the features of property $x$, this prior can be used to reflect the mechanism of the efficient algorithm. 
Specifically, once $d_i \neq 0$, the algorithm no longer allocates budget to that task which creates truncation in the distribution of tasks and information about $x$ is lacking.
If the efficient budget allocation algorithm is performing well, we can reason that the tasks that are considered complete could have a different distribution for $x$ than what $\xhat$, estimated from the truncated crowd data, tells us. 
The priors $D_d$ can reflect our understanding of how $x$ may be affected by the allocation algorithm. 
Next, to develop a single probability distribution for $x$, DEPS continues by generating a sample $\xtilde$ to serve as a debiased $x$ (as opposed to, for example, working with a mixture of $N$ per-task distributions): for each task $i \in [1,\ldots,N]$, sample $\xtilde_i$ from the distribution $P(x_i \mid \{y_{ij}\}, d_i)$. 
Inference on these aggregated $\{\xtilde\}$ is then performed using standard techniques such as maximum likelihood estimation (MLE) or the method of moments. 
For MLE, for example, parameters $\theta$ for $P(\xtilde)$ are determined by maximizing the joint log-likelihood $\ln \mathcal{L}\left(\theta \mid \{\tilde{x}\}\right)$ of the parameters given the data.

\begin{figure*}%
\centering
\subfigure[Less confidence is algorithm's decisions]{%
    \includegraphics[width=.425\textwidth]{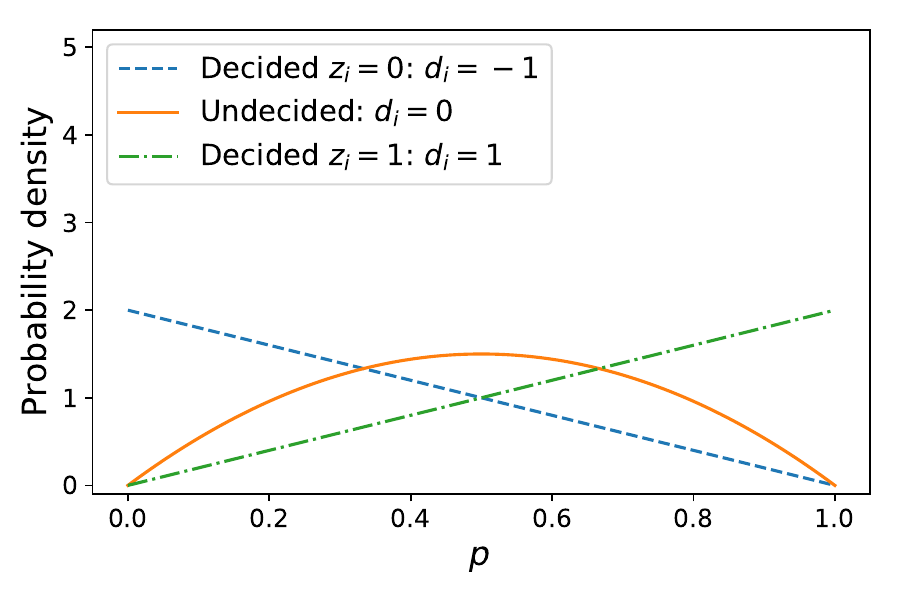}
    \label{fig:decision_priors:2}
}
\subfigure[More confidence in decisions]{%
\includegraphics[width=.425\textwidth]{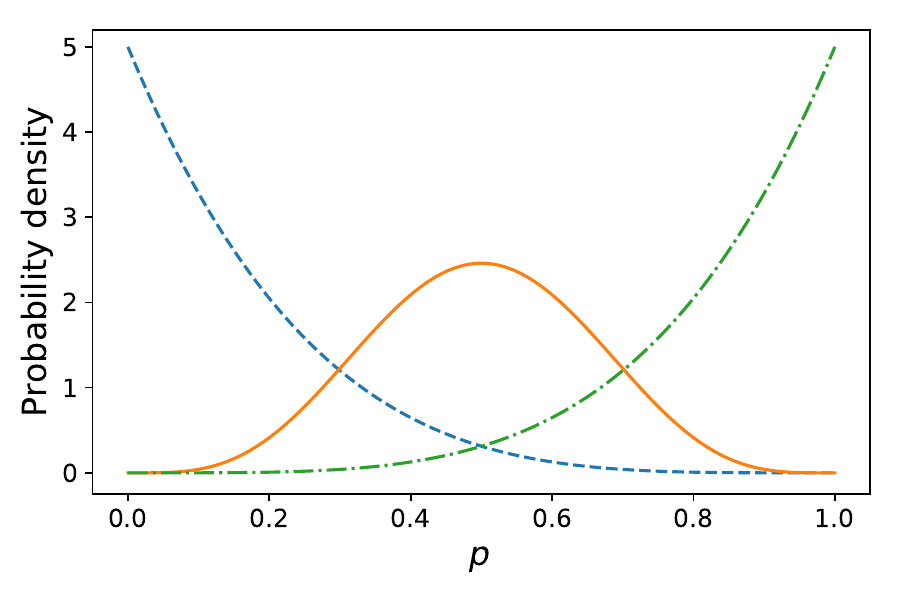}
    \label{fig:decision_priors:5}
}
\caption{Example decision priors for when a researcher is 
\lett{a} less confident in the allocation method's decisions, 
\lett{b} more confident in the decisions.
\label{fig:decision_priors}}
\end{figure*}

As a concrete application of DEPS, we now focus on the property of problem difficulty discussed above (where now $x=p$); we discuss other properties in Sec.~\ref{subsec:otherDEPS}.
An allocation algorithm will introduce bias into the data, making the point estimates $\phat_i = a_i/n_i$ unreliable (Fig.~\ref{fig:req_bias:all_tasks_req}).
For problem difficulty $p$, the beta distribution is the natural choice of prior: $p_i \mid d_i \sim \mathrm{Beta}(\alpha_{d_i}, \beta_{d_i})$ with parameters $\Phi_{d_i} = (\alpha_{d_i}, \beta_{d_i})$.
Worker responses are modeled $y_{ij} \sim \text{Bernoulli}(p_i)$,  then the posterior for $p_i$ (Eq.~\ref{eq:posteior_model}) will also follow a beta distribution with parameters $a_i+\alpha_{d_i}$ and $b_i+\beta_{d_i}$, where $a_i$ and $b_i$ are the number of 1 and 0 labels, respectively, for task $i$. 
The choice of hyperparameters $\alpha_d$ and $\beta_d$ gives the researcher flexibility in how they incorporate the efficient algorithm's decision into the inference model. 
For example, a beta prior that is skewed towards $p=0$ could be used for tasks that have $d_i = -1$ with another beta skewed towards $p=1$ used for $d_i = 1$. 
The amount of skewness of the $d \neq 0$ priors can then reflect how confident the researcher is in the allocation algorithm's decision.
And tasks that are undecided will have priors that provide more weight near $p=1/2$.
We illustrate two choices of decision priors in Fig.~\ref{fig:decision_priors} and we discuss a data-driven approach to calibrating these priors in Sec.~\ref{subsec:calibratingdecisionpriors}.

Having generated a sample $\{\ptilde\}$, inference is performed using either standard maximum likelihood estimators or, as an alternative, method of moments estimators.
For maximum likelihood, the joint log-likelihood for the parameters of (in this case) a beta distribution given the data are
\makeatletter
\if@twocolumn
\begin{equation}
\begin{aligned}
\ln \,{\mathcal{L}}(\alpha ,\beta \mid \{\tilde{p}\}) &= \sum _{i=1}^{N}\ln {\mathcal {L}}_{i}(\alpha ,\beta \mid \tilde{p}_{i})\\
&=(\alpha -1)\sum _{i=1}^{N}\ln \tilde{p}_{i} \\& \quad + (\beta-1)\sum _{i=1}^{N}\ln(1-\tilde{p}_{i})\\& \quad-N\ln \mathrm {B} (\alpha ,\beta ).\end{aligned}
\label{eqn:loglike_beta_MLE}
\end{equation}
\else
\begin{equation}
\begin{aligned}
\ln \,{\mathcal{L}}(\alpha ,\beta \mid \{\tilde{p}\})&=\sum _{i=1}^{N}\ln {\mathcal {L}}_{i}(\alpha ,\beta \mid \tilde{p}_{i})\\
&=(\alpha -1)\sum _{i=1}^{N}\ln \tilde{p}_{i} + (\beta-1)\sum _{i=1}^{N}\ln(1-\tilde{p}_{i})-N\ln \mathrm {B} (\alpha ,\beta ).\end{aligned}
\label{eqn:loglike_beta_MLE}
\end{equation}
\fi
\makeatother
MLE values for $\alpha$ and $\beta$ are then found numerically by solving:
\begin{align} 
\frac {\partial \ln {\mathcal {L}}(\alpha ,\beta \mid \{\tilde{p}\})}{\partial \alpha } &=\sum_{i=1}^{N} \ln \tilde{p}_{i}-N\left(\psi(\alpha)-\psi(\alpha + \beta)\right) =0 \\
\frac {\partial \ln {\mathcal {L}}(\alpha ,\beta \mid \{\tilde{p}\})}{\partial \beta } &= \sum _{i=1}^{N}\ln(1-\tilde{p}_{i})-N\left(\psi (\beta ) -\psi (\alpha +\beta )\right)=0
\label{eq:alpha_beta_hat_MLE}
\end{align}
where $\psi$ is the digamma function.
The method of moments estimators, meanwhile, are:
\begin{align}
\hat{\alpha} = \pline \left(\frac{{\pline}\left(1-{\pline}\right)}{S_{\phat}^{2}}-1\right), \quad
\hat{\beta} = \left(1-\pline\right)\left(\frac{\pline\left(1-\pline\right)}{S_{\phat}^{2}}-1\right), 
\label{eq:methodm_beta}
\end{align}
where $\pline$ and $S_{\phat}^2$ are the sample mean and variance, respectively, of $\phat$.

\subsection{Calibrating decision priors}
\label{subsec:calibratingdecisionpriors}

Our approach to inferring the difficulty distribution introduces decision priors $D_d$ to integrate the crowdsourcing allocation algorithm into the inference.
The exact choice of these priors is up to the needs of the research; this freedom provides the researcher a framework for expressing their confidence in the algorithm.
For example, a researcher studying problem difficulty $p$ who is very confident in the algorithm's decision $d_i = 1$ ($d_i = -1$) would select priors that are sharply peaked at $p_i=1$ ($p_0 = 0$).
Likewise, a researcher with less confidence in the decisions would choose flatter priors over $0 \leq p_i \leq 1$.

Depending on their problem, researchers may have \emph{a priori} reasons for choosing their priors. But, sometimes, they may wish to calibrate their decision priors from data.
Here we describe one calibration method.
Suppose the researcher has access to a small number of gold standard (GS) crowdsourcing tasks, where the true labels $z_i$ for these items are known.
We propose to apply the crowdsourcing algorithm to these gold standard items in order to estimate its accuracy, then calibrate the decision priors accordingly.

Suppose there are $n_0$ GS tasks with true label $z = 0$ and $n_1$ GS tasks with true label $z=1$. 
Let $m_{00}$ be the number of $n_0$ tasks decided by the algorithm to be label zero ($d=-1$)
and let $m_{01}$ be the number of $n_0$ tasks decided to be label one ($d=1$) (with $m_{00} + m_{01} = n_0$). 
Likewise, define $m_{11}$ and $m_{10}$ for the $n_1$ tasks (with $m_{11} + m_{10} = n_1$).
Now, consider the $d=-1$ prior. 
We argue that the prior probability for $p < 1/2$ should be $m_{00} /  n_0$. 
(In practice, these counts $m$ will likely need to be smoothed to avoid zero counts; see below.)
Likewise, for $d=1$, the prior probability for $p>1/2$ should be $m_{11}/n_1$.
For a beta $d=-1$ decision prior, we can thus choose the Beta parameters to satisfy
\begin{equation}
    \frac{m_{00}}{n_0} = \int_0^{1/2} D_{-1}(p \mid a_{-1}, b_{-1}) \, dp,
\end{equation}
where $D_{-1}$ is the beta prior for $d=-1$ with parameters $a_{-1}$ and $b_{-1}$.
This integral is the regularized incomplete Beta function $I_{1/2}(a_{-1}, b_{-1})$. 
Choosing $a_{-1} =1$ reduces this to $I_{1/2}(1, b_{-1}) = 1-(1/2)^{b_{-1}}$, giving a calibrated $b_{-1}$ of
\begin{equation}
    b_{-1} = \log_2\left(\frac{n_0}{m_{01}}\right).
\end{equation}
Likewise, to calibrate the $d=1$ decision prior $D_1 = \mathrm{Beta}(a_1, b_1)$, choose parameters $a_1$ and $b_1$ that satisfy
\begin{equation}
    \frac{m_{11}}{n_1} = 1-\int_{0}^{1/2} D_1(p \mid a_1, b_1) \, dp.
\end{equation}
Again, choosing $b_1 = 1$ for simplicity gives a calibrated $a_1 = \log_2\left(n_{1} / m_{10}\right)$.

These choices of prior parameters $(1,b_{-1})$ and $(a_1,1)$ now serve to calibrate the decision priors given the performance on the GS tasks.
Of course, not all problems have corresponding GS tasks (indeed, we do not use GS tasks in this work), but for those problems where such tasks are available, this method, which is applicable beyond the case of $x=p$ that we focus on, provides a simple means to inform the prior before continuing on to the main set of tasks.

\subsection{Generalizations and other applications of DEPS}
\label{subsec:otherDEPS}

DEPS is a general-purpose approach to inference about problems when using efficient allocation strategies.
Although we focus our study on inferring the distribution of task difficulties,
here we briefly discuss other applications of DEPS.

One application of DEPS is to infer worker completion times~\cite{10.1145/2736277.2741685,10.1145/3301003,8873609}, the time a worker needs to finish a given task.
If workers are consistently given easy tasks only, then we may have a biased representation of how quickly they can complete work. 
We can use DEPS to infer completion times as follows.
Let $\{S_{ij}\}$ be the completion times for tasks $i=1,\dots, N$ and workers $j=1,\dots, M$, e.g., the number of seconds they took to complete the tasks. Distributions of human interevent times are often modeled with a log-normal,  $\ln(S_{ij}) \sim \mathcal{N}(\mu_j, \tau_j)$~\cite{iribarren2009impact}. 
We establish conjugate priors, $\mu_j\mid \tau_j, d \sim \mathcal{N}(\mu_d, \tau_j)$ and $\tau_j \mid d \sim \mathrm{Gamma}(\alpha_d, \beta_d)$, with $\alpha_d, \beta_d>0$. 
The posterior for $\mu_j$ and $\tau_j$ is 
\makeatletter%
\if@twocolumn%
\begin{multline}
    P(\mu_j, \tau_j \mid d_i, \{S_{ij}\}) \\ 
    \propto \prod_{i\in I_j}P(S_{ij}\mid \mu_j, \tau_j, d_{i}) P(\mu_j\mid \tau_j, d_{i})P(\tau_j \mid d_{i}).
\end{multline}
\else
\begin{equation}
    P(\mu_j, \tau_j \mid d_i, \{S_{ij}\}) \propto \prod_{i\in I_j}P(S_{ij}\mid \mu_j, \tau_j, d_{i}) P(\mu_j\mid \tau_j, d_{i})P(\tau_j \mid d_{i}).
\end{equation}
\fi
\makeatother
This model yields an estimated distribution for completion time for each worker. 
To understand the distribution for all workers, the researcher can sample from each distribution and aggregate the samples, $\tilde{\mu}_j$ and $\tilde{\tau}_j$. 
The distribution $P(\tilde{\mu}, \tilde{\tau}) \approx P(\mu, \tau)$, the true, overall distribution of worker completion times. 
Assuming that completion times and task difficulty are related, we can reflect the decision $d_i$ of the algorithm by varying the parameters of the priors for $\mu_j$ and $\tau_j$ given $d_i$.
For example, a normal prior with a larger mean may be appropriate for a task that has been undecided ($d_i = 0$), while decided tasks ($d_i \neq 0$), should have a prior with a smaller mean. 

In addition to worker completion times, DEPS can also consider task completion times.  
Since the tasks are not randomly shown to the workers the data collected on task completion time will also be biased. 
For this application of DEPS, we again model the time to complete tasks, $S_{ij}$, as a log-normal, $\ln (S_{ij}) \sim \mathcal{N}(\mu_i, \tau_i)$ and establish priors $\mu_i \mid \tau_i, d_{i} \sim \mathcal{N}(\mu_{d_i}, \tau_i)$ and $\tau_i \mid d_{i} \sim \mathrm{Gamma}(\alpha_{d_i}, \beta_{d_i})$, with $\alpha_{d_i}, \beta_{d_i}>0$. 
The posterior for $\mu_i$ and $\tau_i$  of task $i$ is 
\makeatletter%
\if@twocolumn%
\begin{multline}
     P(\mu_i, \tau_i \mid d_i, \{S_{ij}\}) \\ 
    \propto \prod_{j \in J_i}P(S_{ij}\mid \mu_i, \tau_i, d_{i}) P(\mu_i\mid \tau_i, d_{i})P(\tau_i \mid d_{i}).
\end{multline}
\else
\begin{equation}
    P(\mu_i, \tau_i \mid d_i, \{S_{ij}\}) \propto \prod_{j \in J_i}P(S_{ij}\mid \mu_i, \tau_i, d_{i}) P(\mu_i\mid \tau_i, d_{i})P(\tau_i \mid d_{i}).
\end{equation}
\fi
\makeatother
The priors are then adjusted to reflect the decisions $d_i$ in a manner similar to the worker completion time model given above, as is the inference of the overall distribution of task completion times.

Even further generalizations of DEPS are possible.
Briefly, suppose a crowdsourcer is interested in understanding the behavioural traces of workers as they complete tasks.
One approach is to use observed worker dynamics to determine an ``embedding'' vector for each worker. 
These vectors can then be used to predict worker features or activities~\cite{welinder2010multidimensional}.
Let $\mathbf{u}_{ij} = \mathbf{x}_{i} + \mathbf{e}_{ij}$ be a $k$-dimensional data vector representing the behavioural trace data for worker $j$ responding to task $i$, where $\mathbf{x}_i$ is the population data for that task and $\mathbf{e}_{ij}$ is the worker-specific deviations from the population.
Then, let $\mathbf{w}_j$ be a $k$-dimensional unit vector for embedding worker $j$.
This embedding vector and its estimator $\mathbf{\hat{w}}_j$, can capture the various contributions of the behavioural trace data to the worker's labeling decision: the worker responds with $y_{ij}=1$ if the projection $\left<\mathbf{u}_{ij}, \mathbf{w}_j \right> \geq \tau_j$, otherwise $y_{ij}=0$.
Inference of this vector can be performed given priors for the parameters (such as $\tau_j$) and $\mathbf{u}_{ij}$, parameterized by a multivariate Gaussian distribution, for example.
The allocation strategy's decisions can then be incorporated by defining suitable decision priors for $\mathbf{u}_{ij}$, $\mathbf{w}_j$, and so forth, conditioned on $d_i$.  

While only brief outlines, the above applications of DEPS indicate that suitable conditioning of statistical models, even complex ones, using the decision variables, has the potential to improve the information attainable about the crowdsourcing problem at hand.

\section{Materials and Methods}
\label{sec:methods}

In this section, we describe the real-world crowdsourcing datasets we analyze (Sec.~\ref{subsec:datasets}), the synthetic crowdsourcing we simulate (Sec.~\ref{subsec:crowdsims}), and how DEPS is applied including the details of the efficient allocation framework we use (Sec.~\ref{subsec:methods:requallo}).
We focus on using DEPS to infer the task difficulty distribution $\Pr(p)$, where $p$ is the probability a worker response $y_{ij}=1$; tasks with $p\approx 1/2$ are difficult in that it takes many worker responses to accurately distinguish if $z=1$ or $z=0$.
See Sec.~\ref{sec:deps} for a general specification of DEPS along with examples for inferring properties other than $p$.
We also describe a traditional ``baseline'' method to compare DEPS (Sec.~\ref{subsec:baselinemethodwald}) to, and we describe our quantitative measures of evaluation for these methods (Sec.~\ref{subsec:evalperform}).
The results of our experiments are presented in Sec.~\ref{sec:results}

\subsection{Datasets}\label{subsec:datasets}
We study three crowdsourcing datasets.
These data were not generated using an efficient allocation algorithm, and so it has become standard practice to evaluate such algorithms with these data~\cite{li2016crowdsourcing,mcandrew2017reply}--- since labels were collected independently, one can use an allocation algorithm to choose what order to reveal labels from the full set of labels, essentially ``rerunning'' the crowdsourcing after the fact.
Following Li, \emph{et al.}~\cite{li2016crowdsourcing}, we only utilize at most $50\%$ of the total responses available so that the allocation algorithm does not ``run out'' of requested labels as it polls the data.

\begin{description}
\item[RTE] Recognizing Textual Entailment (RTE) dataset~\cite{snow2008cheap}. 
Pairs of written statements were taken from the PASCAL RTE-1 data challenge~\cite{dagan2006pascal} and shown to Amazon Mechanical Turk workers who responded whether or not one statement entailed the other.
These data consist of $N=800$ binary tasks each of which received 10 labels from workers, giving a total of $8,000$ responses.
RTE is available at \url{https://sites.google.com/site/nlpannotations/}.

\item[Bluebirds] Identifying Bluebirds dataset~\cite{welinder2010multidimensional}. 
Each task is a photo of a bluebird, either an Indigo Bunting or a Blue
Grosbeak, and the worker is asked if the photo contains an Indigo Bunting. 
There are $N=108$ binary tasks, and 39 responses for each tasks, therefore, a total of $4,212$ responses. 
Bluebirds is available at \url{https://github.com/welinder/cubam.}

\item[Relevance] Identifying Page Relevance dataset~\cite{lease2011overview}
Each task displays a webpage and a given topic, the worker is asked to determine if the webpage is relevant to the given topic. 
This dataset contains $N=2,275$ tasks, with a range of 1 to 10 responses per task. 
There is on average $6.04$ responses per task. In total there are $13,749$ responses.
Relevance is available at \url{https://sites.google.com/site/treccrowd/2011} (Task 2 test set).

\end{description}
When applying DEPS to infer the distributions of task difficulty $p$ for these datasets,  we used beta decision priors with parameters $\Phi_{-1} = (1,2)$, 
$\Phi_{0} = (2,2)$, and 
$\Phi_{1} = (2,1)$.

\subsection{Allocation method}
\label{subsec:methods:requallo}

We used the Requallo allocation algorithm to choose which task labels are received, either from the real datasets (Sec.~\ref{subsec:datasets}) or the simulated crowdsourcing (Sec.~\ref{subsec:crowdsims}).
We chose for completeness a ratio requirement with $c=4$ (Sec.~\ref{subsec:allocalgs}). 
All other details were as those given by Li \emph{et al.}~\cite{li2016crowdsourcing}.
While other allocation methods and parameter choices are worth exploring (see also Discussion), here this method and parameter choices were held fixed so that post-allocation inference methods DEPS and the baseline method (see below) are always compared on the same collected data.

\subsection{Baseline method---Wald estimation}
\label{subsec:baselinemethodwald}

To understand the performance of DEPS, we compare to the following baseline method, known as Wald estimation, which is a conventional approach to this inference problem.
Wald estimation uses the MLE estimator for $p$: for each task $i$, $\phat_i = a_i / n_i$, where $a_i$ is the number of positive responses to $i$ from workers and $n_i$ is the total number of responses from workers.
The $\phat$ are then used directly to approximate the distribution $\Pr(p)\approx \Pr(\phat)$ and estimate the beta parameters using Eqs.~\eqref{eqn:loglike_beta_MLE}--\eqref{eq:alpha_beta_hat_MLE} or Eq.~\eqref{eq:methodm_beta}.

Unlike DEPS (Sec.~\ref{sec:deps}), this baseline estimation using $\phat$ suffers from small-data problems.
Specifically, if either $a_i=0$ or $a_i = n_i$, then the likelihood used in MLE is undefined.
Yet, either situation is likely when only a few labels are collected for task, which can often occur when using an efficient allocation strategy (see Fig.~\ref{fig:req_bias:all_tasks_req}).
A common solution to this problem is to use a \emph{smoothed} estimate of $\phat$, where $a_i$, $b_i$, and $n_i$ are replaced with $a_i+\epsilon$, $b_i +\epsilon$, and $n_i + 2\epsilon$, respectively (we use $\epsilon=1$). 
Unfortunately, while smoothing to this degree using ``pseudo-labels'' is common, such a level of smoothing may overly bias our Wald estimates of $\phat$ towards $\phat=1/2$, so we also explore an alternative solution:
the values of $\phat$ are \emph{transformed} using $\left(\phat(N-1) +1/2 \right)/N$ such that now $\phat \in (0,1)$ instead of $\phat \in [0,1]$~\cite{smithson2006better}.
For these two procedures, when fitting to $\{\phat\}$, we found better parameter estimates when using MLE for smoothed Wald and Method of Moments for transformed Wald. 
DEPS, in contrast, is not affected by this small-data problem.

\subsection{Crowdsourcing simulations}
\label{subsec:crowdsims}

We wish to supplement our results using real crowdsourcing with controlled simulations of crowdsourcing problems. 
To generate synthetic datasets according to the crowdsourcing model defined in Sec. \ref{sec:background}, we assume each task $i$ has an intrinsic parameter $p_i$ with each worker response to task $i$ as a Bernoulli variable with parameter $p_i$.
This probability governs how difficult a task is in terms of how many responses are necessary to determine its label: the closer $p$ is to $1/2$ the more labels are necessary to accurately distinguish $z=0$ from $z=1$.
We further endow the model with a prior probability distribution on $p_i$, specifically $p \sim \mathrm{Beta}(\alpha, \beta)$, a beta distribution with hyperparameters $\alpha$ and $\beta$.
This prior distribution lets us generate $N$ tasks and control the difficulty of each by determining how many tasks are near $p\approx 1/2$ and how many tasks are near $p \approx 0$ or $p \approx 1$.
Conversely, statistical inference can be performed to determine the posterior distribution of $p$ given the data $\{y_{ij}\}$.
Using this simulation model we can implement efficient budget allocation techniques such as Requallo in order to study the effect of efficient allocation on the distribution $\Pr(p)$. 
Unlike with the real data, the true distribution of $p$ is known, and we can test inference methods by comparing their estimates to the true distribution.

To apply DEPS to simulated data, we used decision prior parameters 
$\Phi_{-1} = (1,5)$, 
$\Phi_{0} = (5,5)$, and 
$\Phi_{1} = (5,1)$. 
These are more confident priors than the ones used with real data (Sec.~\ref{subsec:datasets}); see Sec.~\ref{sec:discussion} for further discussion.

\subsection{Evaluating performance}
\label{subsec:evalperform}

We use an information-theoretic measure to quantitatively compare the distribution $\Pr(p)$ of task difficulty $p$ found under various conditions.
For synthetic datasets we have imposed the ground truth distribution of $p$, so we can compare our inferred estimates to this known ground truth. 
The ground truth distributions are not available to us when examining the real datasets, so instead we compare the estimated $\Pr(p)$ found with a biased portion of the data revealed using Requallo with the estimated $\Pr(p)$ found using all the data, which were collected in an unbiased manner.

The Kullback-Leibler (KL) divergence (or relative entropy) between two random variables $X$ and $Y$ (measured in `nats') is 
\makeatletter%
\if@twocolumn%
\begin{align} D _ { \mathrm { KL } } \left( X, Y \right) & = \int_{-\infty}^{\infty} f_X ( x ; \theta ) \ln \left( \frac { f_X ( x ; \theta ) } { f_Y \left( x ; \theta^\prime \right) } \right) dx \label{eqn:KLdivGeneral} \\ 
\begin{split}%
& = \ln \left(\frac{\mathrm{B}\left(\alpha^{\prime},\beta^{\prime} \right)}{\mathrm{B}\left( \alpha , \beta \right)} \right) + \left( \alpha - \alpha ^ { \prime } \right) \psi ( \alpha ) \\
& \quad + \left( \beta - \beta^{\prime} \right) \psi(\beta) \\ & \quad+ \left( \alpha^{\prime} - \alpha + \beta^{\prime} - \beta\right) \psi(\alpha + \beta),
\label{eq:DKLforBetas}
\end{split}
\end{align}
\else
\begin{align} D _ { \mathrm { KL } } \left( X, Y \right) & = \int_{-\infty}^{\infty} f_X ( x ; \theta ) \ln \left( \frac { f_X ( x ; \theta ) } { f_Y \left( x ; \theta^\prime \right) } \right) dx \label{eqn:KLdivGeneral}\\ 
& = \ln \left(\frac{\mathrm{B}\left(\alpha^{\prime},\beta^{\prime} \right)}{\mathrm{B}\left( \alpha , \beta \right)} \right) + \left( \alpha - \alpha ^ { \prime } \right) \psi ( \alpha ) + \left( \beta - \beta^{\prime} \right) \psi(\beta) + \left( \alpha^{\prime} - \alpha + \beta^{\prime} - \beta\right) \psi(\alpha + \beta), 
\label{eq:DKLforBetas}
\end{align}
\fi
\makeatother
where $f_X$ ($f_Y$) is the density for $X$ ($Y$), and the second line holds for the case where $X$ and $Y$ are both beta-distributed: 
$X \sim \mathrm{Beta}\left(\alpha,\beta\right)$ and $Y \sim \mathrm{Beta}\left(\alpha^\prime,\beta^\prime\right)$,
$\mathrm{B}$ is the Beta function, and $\psi$ is the digamma function.
We utilize Eq.~\eqref{eq:DKLforBetas} to measure how well our inference method estimates the underlying prior beta distribution of $p$.
If the distributions of $p$ are not Beta, or one is interested in a different property than $p$, one can still use the general Eq.~\eqref{eqn:KLdivGeneral}, perhaps with an appropriate sample estimator for the KL-divergence~\cite{perez2008kullback}.
We take $X$ to be the ground truth distribution for $p$ (synthetic data) or the distribution  estimated using the full, unbiased data (real data), and $Y$ to be the distribution inferred using DEPS on Requallo-collected data.

\section{Experiments}
\label{sec:results}

We divide our experiments into those analyzing real-world data (Sec.~\ref{subsec:realworldresults}, Figs.~\ref{fig:PDF_dataDistributions}--\ref{fig:timeseries}, and Table~\ref{tab:realworldalphabeta}) and those analyzing synthetic crowdsourcing simulations (Sec.~\ref{subsec:syntheticresults} and Fig.~\ref{fig:kldiv_heatmap}). %

\subsection{Real-world data}
\label{subsec:realworldresults}

\begin{figure*}
\centering
\subfigure[RTE]{
    \includegraphics[width=.31\textwidth]{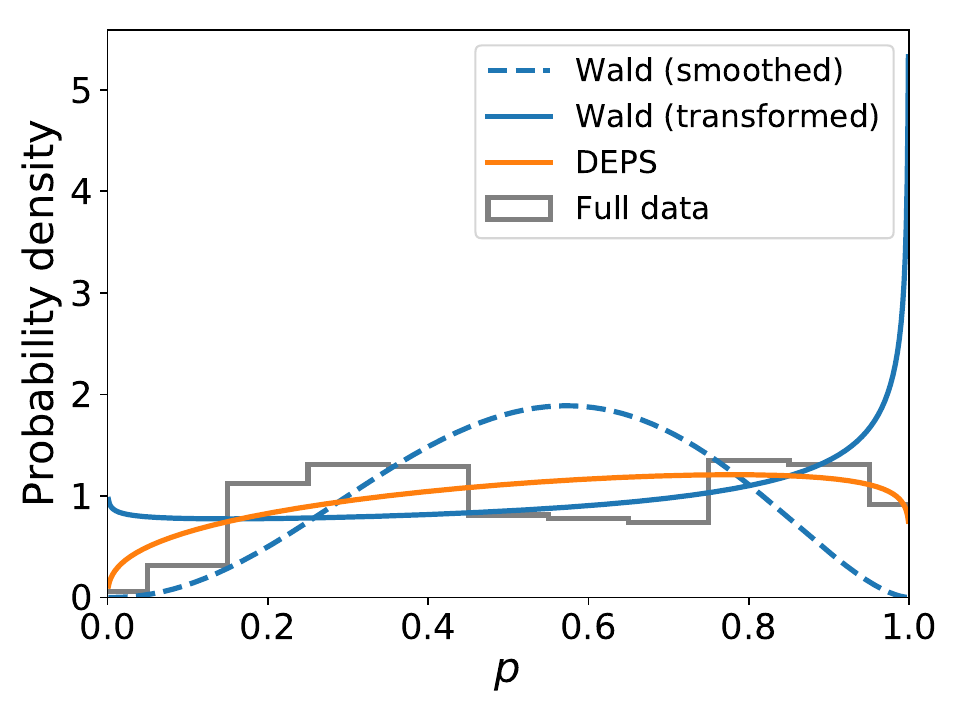}
    \label{fig:PDF_dataDistributions:RTE}
}
\subfigure[Bluebirds]{
\includegraphics[width=.31\textwidth]{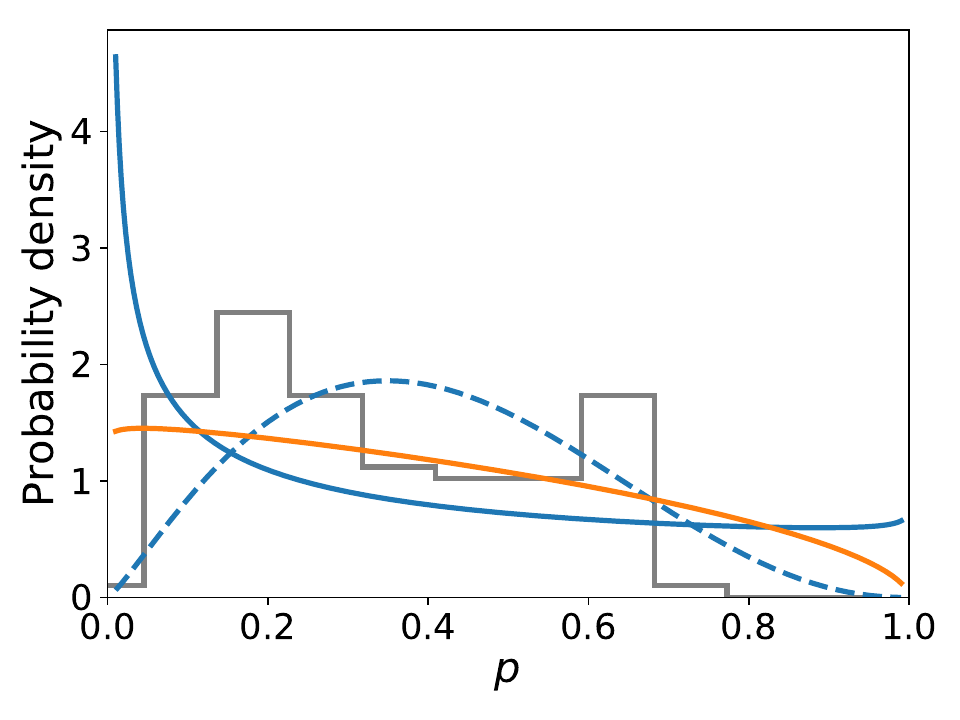}
    \label{fig:PDF_dataDistributions:BB}
}
\subfigure[Relevance]{
\includegraphics[width=.31\textwidth]{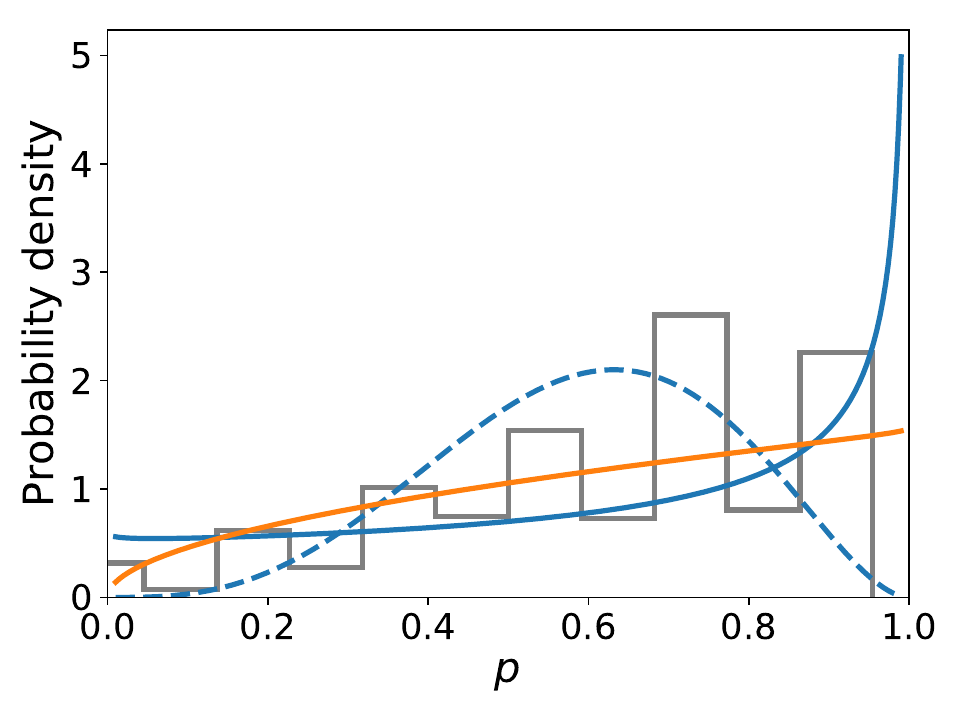}
    \label{fig:PDF_dataDistributions:REL}
}
\caption{%
Inference of problem difficulty distributions using 25\% of labels.
Tasks are allocated using Requallo, then Wald estimation or DEPS estimation of problem difficulty is performed. 
These results are compared to the distribution of $\phat$ estimated using the full dataset (100\% of labels), show as histograms.
The ``smoothed'' and ``transformed'' variants of Wald estimation tend to add too much probability near either the center ($p=1/2$) or extremes ($p=0,1$).
DEPS achieves good agreement with the full data distribution.
\label{fig:PDF_dataDistributions}
}
\end{figure*}

Figure~\ref{fig:PDF_dataDistributions} considers estimation of the difficulty distribution $\Pr(p)$ when limited to 25\% of the available budget. 
For the Wald baseline method, Fig.~\ref{fig:PDF_dataDistributions} show how efficient allocation biases the estimated $\phat$ towards $1/2$ in the "smoothed" case and toward $0$ or $1$ in the "transformed" case.
In contrast to Wald, which requires either smoothed or transformed point estimates,
DEPS shows good qualitative agreement with the full data, and fits the unmodified data, making it more robust than Wald to the bias from efficient allocation. 

\begin{figure*}
    \centering
    \includegraphics[width=\textwidth]{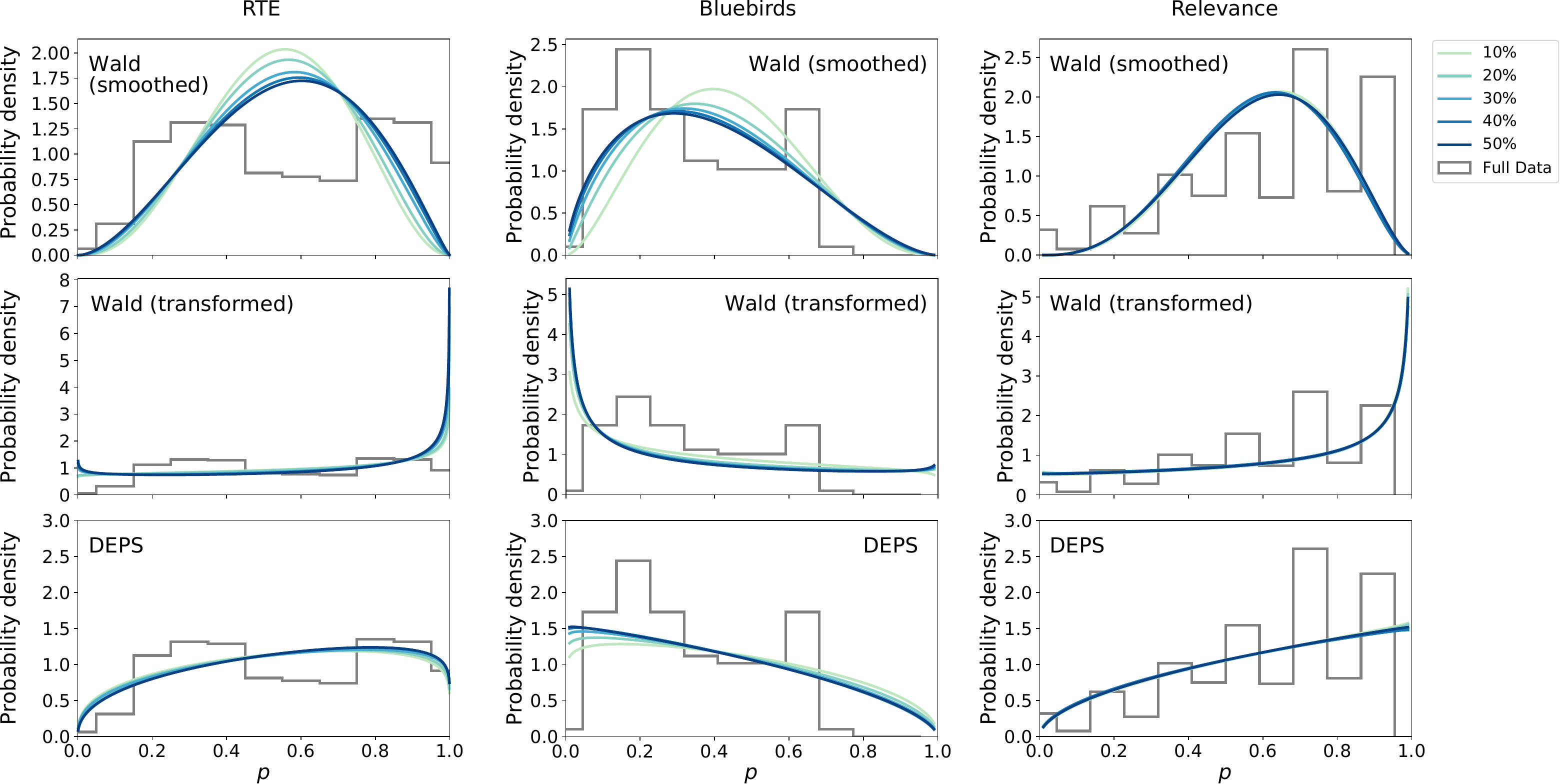}
\caption{Convergence of estimated distributions as more crowd data are used.
Visually, we see that DEPS and Wald (transformed) both converge relatively quickly, often after using only 10\% of the available data.
\label{fig:DEPSrealDataVaryBudget}
}
\end{figure*}

Expanding upon Fig.~\ref{fig:PDF_dataDistributions}
we now investigate in Fig.~\ref{fig:DEPSrealDataVaryBudget} how the estimates of $\Pr(p)$ change as more budget is made available to the crowdsourcer, up to $50\%$ of the total number of labels provided by the data.
We observe in both Wald and DEPS that the estimated distributions converge quickly, often with as little as 10\% of available data.

\begin{figure*}
    \centering
    \subfigure[RTE]{
    \includegraphics[width=.31\textwidth]{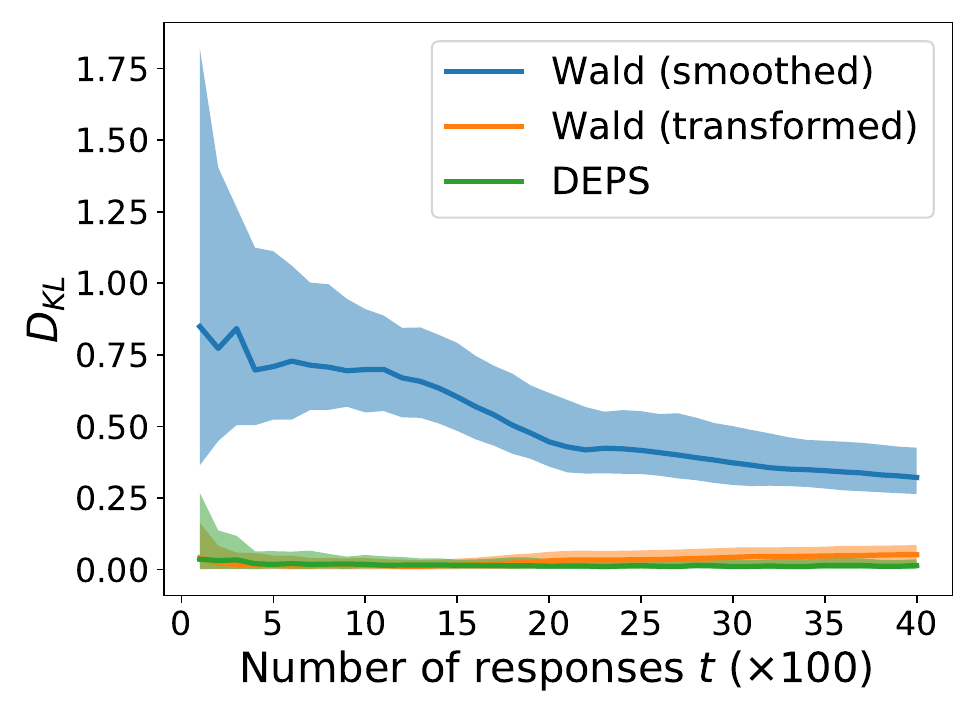}
    \label{fig:timeseries:RTE}
}
\subfigure[Bluebirds]{
\includegraphics[width=.31\textwidth]{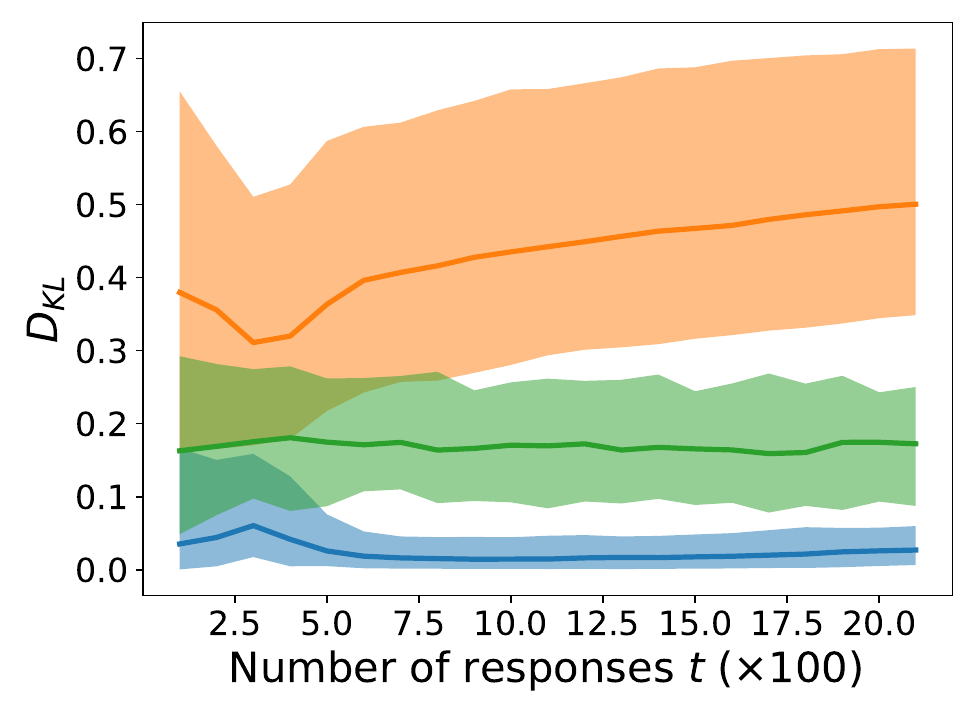}
    \label{fig:timeseries:BB}
}
\subfigure[Relevance]{
\includegraphics[width=.31\textwidth]{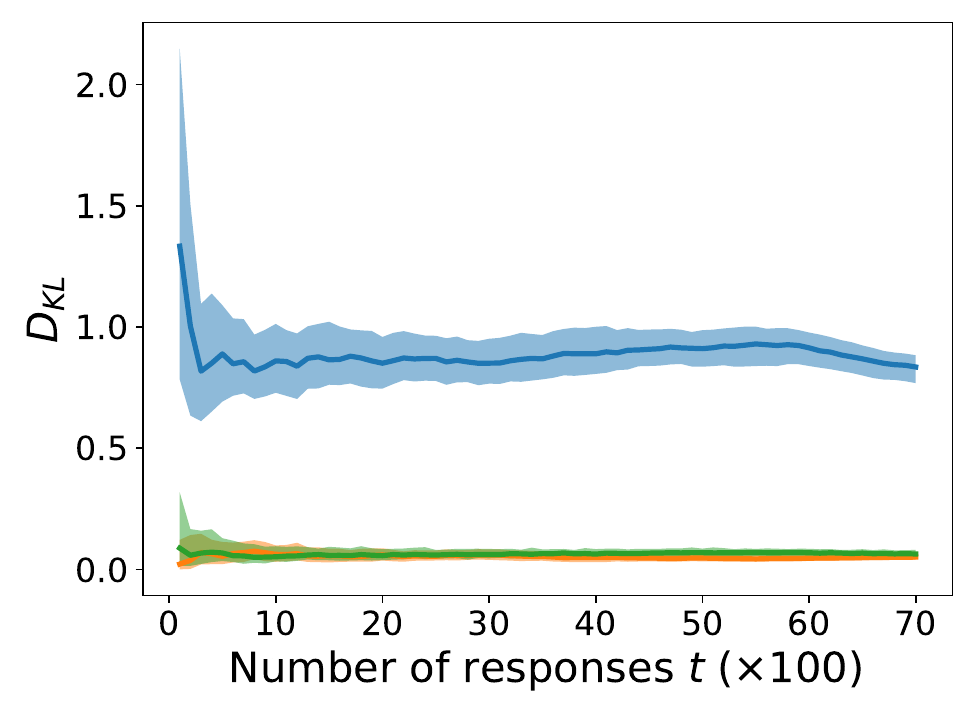}
    \label{fig:timeseries:REL}
}
\caption{
DEPS generally provides more information (lower KL-divergence) about the distribution estimated using the full, unbiased dataset than Wald, whose performance degrades as more (biased) data are received, with the exception of Bluebirds. Shaded areas denote 95\% CI computed over Requallo realizations.
\label{fig:timeseries}
}
\end{figure*}

Expanding on the distribution convergence, Fig.~\ref{fig:timeseries} shows how much information about the distribution of $p$ given by the full (unbiased) data is provided by DEPS and Wald as more biased data are made using Requallo.
With the exception of the Relevance dataset, where performance is relatively comparable, DEPS provides more information as evidenced by the lower KL-divergence (Sec.~\ref{subsec:evalperform}).
Interestingly, Wald shows performance that \emph{degrades} slightly with more data in Fig.~\ref{fig:timeseries:RTE} and \ref{fig:timeseries:BB}, likely due to the fact that, while more data are available, there is more biased data available, as these responses are gathered in a non-uniform manner due to Requallo.
DEPS, however, does not exhibit this degraded performance on these data.
We discuss this further in Sec.~\ref{sec:discussion}.

Lastly, to further compare DEPS and Wald estimates on the real-world datasets, Table~\ref{tab:realworldalphabeta} shows the estimated parameters $\theta = (\alpha, \beta)$  for $\Pr(p)$. 
We applied both methods to 50\% of the data gathered using Requallo (and the bias entailed by Requallo).
For comparison, we also report $\theta$ as estimated using the full, unbiased data.
Although there remains room for improvement, Table~\ref{tab:realworldalphabeta} shows that DEPS achieves estimates of $\theta$ closer to estimates from the unbiased data than Wald does in all cases.

\begin{table*}
\centering\small
\caption{ Parameter estimates using a biased sample compared to parameters inferred using the full, unbiased dataset. 
Bold denotes sample estimates that are closest to estimates from the full data.
 \label{tab:realworldalphabeta}}
\begin{tabular}{llllll}
\toprule
          &        & \multicolumn{2}{c}{Estimates from biased $50\%$ data} & \multicolumn{2}{c}{Full data} \\
Dataset   & Method & \multicolumn{1}{c}{$\alpha$ [95\% CI]} & \multicolumn{1}{c}{$\beta$ [95\% CI]} & \multicolumn{1}{c}{$\alpha$} & \multicolumn{1}{c}{$\beta$} \\
\midrule
RTE       & Wald (smoothed)      & $3.04~[2.83,3.35]$     & $2.41~[2.28,2.63]$     &  $1.24$              & $0.92$   \\
          & Wald (transformed)   & $0.88~[0.78,1.05]$     & $0.62~[0.53,0.73]$     &  $1.24$              & $0.92$   \\
          & \textbf{DEPS}                 & $\textbf{1.43}~[1.27, 1.63]$    & $\textbf{1.12}~[0.97,1.27]$     &  $1.24$              & $0.92$   \\[0.5em]
Bluebirds & \textbf{Wald} (smoothed)      & $\textbf{2.02}~[1.82, 2.30]$    & $\textbf{3.09}~[2.65,3.69]$     &  $2.25$              & $3.66$   \\
          & Wald (transformed)   & $0.46~[0.31, 0.61]$    & $0.88~[0.63,1.19]$     &  $2.25$              & $3.66$   \\
          & DEPS                 & $1.00~[0.78, 1.25]$    & $1.56~[1.28,2.07]$     &  $2.25$              & $3.66$   \\[0.5em]
Relevance & Wald (smoothed)      & $4.13~[3.98, 4.26]$    & $2.79~[2.70,2.86]$     &  $1.51$              & $0.72$   \\
          & Wald (transformed)   & $0.98~[0.93,1.03]$     & $0.49~[0.46,0.52]$     &  $1.51$              & $0.72$   \\
          & \textbf{DEPS}                 & $\textbf{1.53}~[1.53, 1.61]$    & $\textbf{1.00}~[0.95,1.06]$     &  $1.51$              & $0.72$   \\ \bottomrule
\end{tabular}
\end{table*}

\subsection{Synthetic data}
\label{subsec:syntheticresults}

\begin{figure}[h]
\centering
\includegraphics[width=0.6\textwidth]{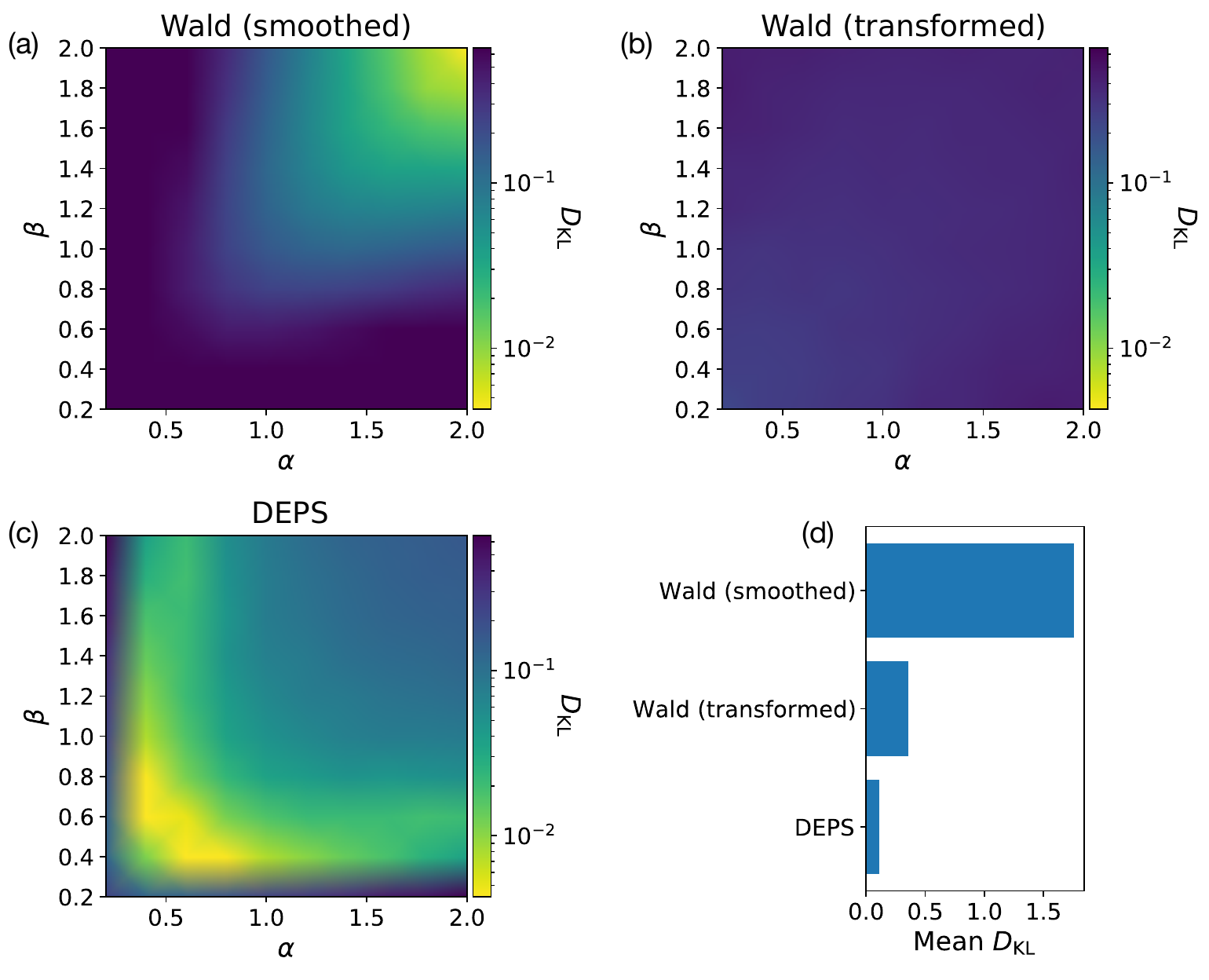}
\caption{The KL-divergence (or relative entropy) $D_\mathrm{KL}$ between the inferred and true distributions of $p$ over a range of parameters for the true distribution.
Here we see that DEPS outperforms the Wald baselines for most parameter values, achieving a lower divergence in its estimates of the true distribution.
Panel d shows the mean KL-divergence averaged over the values of $\alpha$ and $\beta$ in the matrices shown in panels a-c.
\label{fig:kldiv_heatmap}
}%
\end{figure}

Supplementing our results using real datasets, we also explored the performance of DEPS and the Wald baseline method for crowdsourcing problems where the true distribution $\Pr(p)$ is known.
Figure \ref{fig:kldiv_heatmap} shows the KL-divergence $D_\mathrm{KL}$ between estimated $\Pr(p)$ and the true distribution across a range of parameter values ($\alpha,\beta$).
Across most parameters, except for some cases with extremely small values of $\alpha$ or $\beta$, DEPS provides more information (lower $D_\mathrm{KL}$) than Wald does about the true distribution. (Note the logarithmic color scale for $D_\mathrm{KL}$ used in Fig.~\ref{fig:kldiv_heatmap}(a)-(c).)
Taken together, DEPS is able to generate more information about the underlying distribution $\Pr(p)$ than baseline methods even when baseline methods use the same efficient allocation algorithm.

\section{Discussion}
\label{sec:discussion}

Efficient crowd allocation methods such as Requallo can minimize the costs of crowdsourcing, but we showed that this efficiency comes at a cost: the data collected are now biased. 
Indeed, in our results we showed that Requallo completed far more easy tasks and far fewer hard tasks, in terms of reaching a consensus from as few responses as possible.
For some tasks this bias may not be a concern, but many tasks, such as studying the behavior of the crowd as they perform the work, may be harmed by this bias. 
Even the common task of using crowdsourcing to gather machine learning training data may be hindered by an allocation method: images, for example, where the crowd most disagrees about the label are some of the most useful examples when training an image classifier, and so if anything more data should be collected for those images, not less.

Our experiments (Sec.~\ref{sec:results}) provide evidence supporting the performance of DEPS, both its accuracy and its efficiency, but more work is warranted in several directions.
While we generally found that DEPS provided the best performance, this was not always the case:
smoothed Wald performed best on the Bluebirds data and so it is worth exploring which methods are most appropriate to which data and when.
Likewise, we focused our validation procedure around a single inference task, estimating the difficulty distribution for a crowdsourcing problem, and a single allocation method, Requallo.
Future work should consider other problem properties, such as better understanding completion times or behavioural traces of works, as we briefly sketched out in Sec.~\ref{subsec:otherDEPS}, as well as better exploring how well DEPS can work with allocation methods besides Requallo and how well DEPS compares with (or can be combined with) other inferential methods~\cite{10.1145/3132847.3133002,9284535}.
DEPS also works using an informative ``decision'' prior, and this introduces subjectivity into the property inference.
We provided some guidance for calibrating decision priors, but there remains considerable researcher flexibility.
Indeed, we found variations in DEPS performance across the parameter space in Fig.~\ref{fig:kldiv_heatmap} that imply care may be needed when matching decision priors to specific problems.
A systematic study of choosing and tuning decision priors efficiently is warranted.

It is possible that the performance of DEPS can degrade as more data are collected.
As more labels are received overall, more difficult tasks are more likely to reach the completeness requirement, pushing the ``horizon'' of completeness towards $p = 1/2$. 
In this case, we may want to flatten our priors accordingly. %
In other words, we motivated the choice of steep decision priors in Sec.~\ref{sec:deps} as representing the confidence that we have in the algorithm's decisions, but, in fact, that steepness is also related to the budget available to the crowdsourcer. 
Incorporating a budget dependency into the design of the decision priors is therefore an important avenue for improvement, particularly for a large-scale deployment of DEPS.
This would be especially interesting for crowdsourcing problems designed to distribute the crowd non-uniformly, deploying more workers in some areas of the problem space than others and adapting (perhaps dynamically) the DEPS decision priors accordingly.

\section{Conclusion}
\label{sec:conclusion}

In this work, we introduced Decision Explicit Probability Sampling (DEPS), a flexible approach to estimating a distribution when given biased data. 
By explicitly incorporating an allocation method's decisions into prior distributions, DEPS can adjust for the bias induced by the allocation method. 
Using Requallo as an exemplar allocation strategy, DEPS estimates a more accurate distribution for the true, unknown property distribution than if estimation were performed using traditional methods. 
This allows researchers to still collect data from the crowd efficiently, while being able to extract better information about more than just the correct label for a task.

In summary, DEPS provides a method to account for the bias introduced by efficient allocation algorithms in order to better understand properties of interest for a crowdsourcing problem.
The researcher can apply DEPS to problems of interest by adapting the decision priors to reflect how the allocation method introduces bias. 
The flexibility of DEPS allows for researchers to implement DEPS for a variety of research questions, such as understanding problem difficulty, worker completion times, and behavioral traces.
Researchers can implement efficient crowdsourcing while performing accurate inference about the distribution of interest, showing that DEPS can help address one of the key challenges of crowdsourcing: maximizing the information gained from finite, and often costly to gather, data.

\section*{Acknowledgments}

This material is based upon work supported by the National Science Foundation under Grant No.\ IIS-1447634.

\singlespacing
\bibliographystyle{ieeetr}
\bibliography{main}{}

\end{document}